\pdfoutput=1
\PassOptionsToPackage{prologue,dvipsnames}{xcolor}

\documentclass[11pt]{article}

\usepackage[]{acl}

\usepackage{times}
\usepackage{latexsym}

\usepackage[T1]{fontenc}
\usepackage[utf8]{inputenc}

\usepackage{microtype}

\usepackage{inconsolata}

\usepackage{graphicx}
\usepackage{booktabs}
\usepackage{multirow}
\usepackage{multicol}
\usepackage{xcolor}
\usepackage{pifont}
\usepackage{amsmath}
\usepackage{mdframed}   % 박스 스타일 추가
\usepackage{enumitem}   % 리스트 스타일 조정
\usepackage{colortbl}

\title{Teach-to-Reason with Scoring:\\
Self-Explainable Rationale-Driven Multi-Trait Essay Scoring}

\author{Heejin Do$^{1}$, Sangwon Ryu$^{1}$, Gary Geunbae Lee$^{1,2}$ \\
  \centering
  \begin{tabular}[t]{c}
    $^{1}$Graduate School of Artificial Intelligence, POSTECH, Republic of Korea \\
    $^{2}$Department of Computer Science and Engineering, POSTECH, Republic of Korea \\
    \texttt{\{heejindo, ryusangwon, gblee\}@postech.ac.kr} \\
  \end{tabular}
}
\begin{document}
\maketitle

\definecolor{sky}{RGB}{204, 229, 255}
\definecolor{org}{RGB}{255, 229, 204}
\definecolor{gre}{RGB}{255, 255, 229}
\definecolor{pur}{RGB}{255,204,255}
\definecolor{pin}{RGB}{255,204,229}

\begin{abstract}

Multi-trait automated essay scoring (AES) systems provide a fine-grained evaluation of an essay's diverse aspects. While they excel in scoring, prior systems fail to explain why specific trait scores are assigned. This lack of transparency leaves instructors and learners unconvinced of the AES outputs, hindering their practical use. To address this, we propose a self-explainable Rationale-Driven Multi-trait automated Essay scoring (RaDME)\footnote{Codes and all generated results will be publicly available.} framework. RaDME leverages the reasoning capabilities of large language models (LLMs) by distilling them into a smaller yet effective scorer. This more manageable \textit{student} model is optimized to sequentially generate a trait score followed by the corresponding rationale, thereby inherently learning to select a more justifiable score by considering the subsequent rationale during training. %effectively harnessing the explanatory power extracted from the \textit{teacher} LLM. 
%By training the model to consider the subsequent "reason" during score prediction, there is a likelihood of indirectly learning from future tokens, thus leading the model to select a more justifiable score.
Our findings indicate that while LLMs underperform in direct AES tasks, they excel in rationale generation when provided with precise numerical scores. Thus, RaDME integrates the superior reasoning capacities of LLMs into the robust scoring accuracy of an optimized smaller model. Extensive experiments demonstrate that RaDME achieves both accurate and adequate reasoning while supporting high-quality multi-trait scoring, significantly enhancing the transparency of AES.

\end{abstract}
\section{Introduction}
Fine-grained feedback, grounded in an accurate assessment of writing quality, is crucial for enhancing learners' writing skills. While traditional holistic automated essay scoring (AES) models \cite{taghipour2016neural, dong2016automatic, dong2017attention, wang2022use} provide only an overall score, recent research has shifted toward multi-trait scoring \cite{kumar2022many, do-etal-2024-autoregressive, do-etal-2024-autoregressive-multi} to enable a more granular evaluation of essays. With the introduction of the autoregressive score generation framework, ArTS \cite{do-etal-2024-autoregressive}, multi-trait scoring has made remarkable strides, achieving substantial agreement with human-expert ratings.

%Unlike earlier holistic AES \cite{taghipour2016neural, dong2016automatic, dong2017attention, wang2022use} that provide only an overall score, multi-trait assessment offers a more comprehensive understanding of essay quality. 

Despite advancements in AES, current systems remain opaque, as they fail to explain the rationale behind their scoring decisions. While these models deliver accurate score predictions, their lack of interpretability undermines the transparency and reliability of assessments \cite{kumar2020explainable, johnson2024examining}. Thus, educators and students, who require more than just numerical feedback, often find these outputs unconvincing, restricting the practical deployment of AES. %require clear insights into why certain trait scores are given rather than merely receiving numerical feedback. 

 %Educators and students require actionable insights into why certain trait scores are given rather than merely receiving numerical feedback. 
%For instance, students may ask, ``Why did my essay receive a Content score of 7?''—but existing systems overlook this challenge. Ensuring transparency in scoring decisions is essential for AES to achieve the reliability needed for practical educational deployment.

\begin{figure}[t]
\centering
\includegraphics[width=1\linewidth]{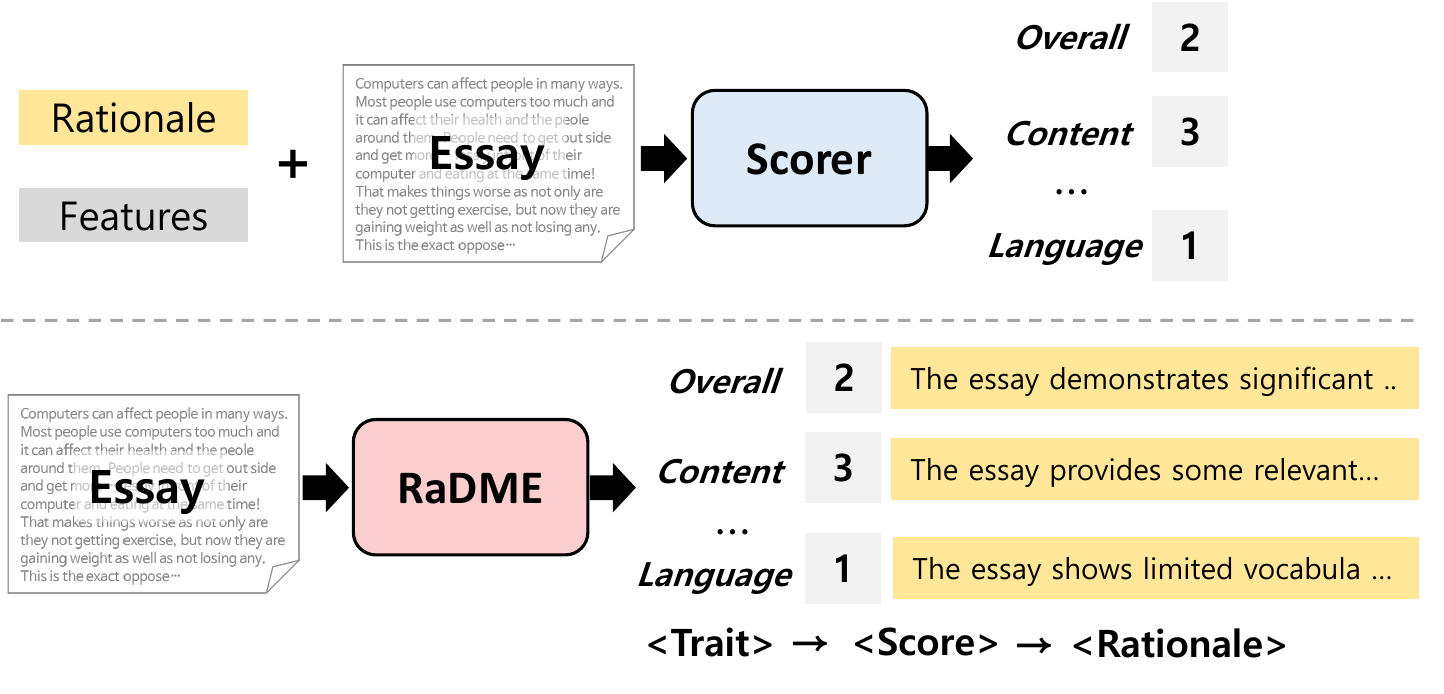}
\caption{Comparison of existing multi-trait scoring methods (top) and RaDME (bottom). While existing methods take features or rationales as input, not allowing direct interpretation of the results; however, RaDME explicitly derives scores followed by its rationales, enhancing the reliability of the outcomes.}
\label{fig1}
\end{figure}

To interpret the model decisions, prior studies have attempted to derive scoring decisions by leveraging explicit grammatical or linguistic features \cite{wang2021prompt, sudoh2024automated}. However, these approaches focus on model-driven explanations rather than providing human-centered justifications for assigned scores. 
%however, their explainability is limited to the model’s perspective, failing to provide human-centered explanations for understanding the assigned scores. 
More recently, \citet{chu2024rationale} utilized rubric guidelines to prompt large language models (LLMs) to generate evaluation rationales, which were then used as additional input for an ArTS-based \cite{do-etal-2024-autoregressive} model. While they integrate rationales, their primary goal is to improve AES performance rather than enhance explainability, leaving the scoring model lacking an inherent mechanism to clarify the reasoning behind derived scores (Figure~\ref{fig1}). %Actually, despite the recent surge in attempts to use LLMs for essay scoring, even iterative or sophisticated prompting has failed to achieve accurate scoring \cite{mizumoto2023exploring, mansour2024can, lee-etal-2024-unleashing}. Instead of directly relying on LLMs for scoring, we employ distillation to transfer their reasoning capacity into a more efficient yet effective scoring model.  %Instead, we train the model to describe the reasoning for assessment while deriving the final score.

To address these, we propose a self-explainable, rationale-driven multi-trait essay scoring (RaDME) method, which \textit{learns to reason with scoring}. Drawing inspiration from the human decision-making process, e.g., decide-with-reason,
RaDME is designed to jointly generate a score and its corresponding rationale, ensuring each scoring decision is inherently grounded in clear, justifiable reasoning.
%we have meticulously designed the model to concurrently assess a score and articulate its substantiated rationale, aiming to ensure each scoring decision is deeply rooted in clear reasoning. 
%\texttt{<Trait>$\rightarrow$<Rationale>$\rightarrow$<Score>}.
To achieve this goal, we distill the reasoning capacity of LLMs into a smaller yet effective scoring model. Notably, while LLMs have struggled to achieve precise AES performance even with iterative or sophisticated prompting \cite{mizumoto2023exploring, mansour2024can, lee-etal-2024-unleashing}, we find that they excel at reasoning, particularly when provided with explicit numeric scores; this also aligns with existing research \cite{huang-chang-2023-towards, ryu2024guide}. In contrast, smaller domain-specific expert models excel in scoring but lack reasoning capabilities. RaDME bridges this gap by introducing rationale distillation that maximizes both advantages, allowing for effective rationale-driven assessment. Note that we construct a unified model capable of both reasoning and scoring across multiple traits and prompts by optimizing the model with trait-wise rationale-score pairs as a multi-task learning approach.

%RaDME allows the scoring model to generate reasoning that it previously lacked, thereby enhancing transparency in the final scores. As our goal is to construct a unified model capable of both reasoning and scoring across multiple traits and prompts, we structured the \textit{rationale-score} sequence in a trait-wise manner and trained the model using a multi-task learning approach. 

% likely due to the need for scoring that aligns with specific educational contexts and conditions

% By combining LLMs, which excel at reasoning but struggle with scoring, with small expert models, which are highly accurate in scoring but lack reasoning abilities, RaDME maximizes the strengths of both for optimal evaluation.  

Extensive experimental results demonstrate that RaDME achieves outstanding scoring performance, even surpassing recent state-of-the-art methods while simultaneously generating high-quality rationales. 
%Extensive experimental results demonstrate that RaDME can achieve remarkable scoring performance, outperforming recent robust methods, even while generating rationales. 
This result is particularly noteworthy, as previous attempts to jointly perform feedback generation and scoring have largely failed to achieve reliable scoring \cite{stahl2024exploring}. 
%Detailed investigation of the generated rationales highlights the efficacy of rationale distillation, indicating scoring-first and subsequent reasoning leads to high-quality generation of both. 
%when evaluating the winning rate for rationale generation, the distilled student model closely matches or even surpasses the teacher LLM in relevance and accuracy, highlighting the effectiveness of rationale distillation. 
Further discussions and analyses on both scoring and rationale generation results strongly support RaDME’s ability to enhance both reasoning capabilities and scoring quality. Our findings include that RaDME with scoring-first and subsequent reasoning notably enhances both generations. Our work takes a crucial step toward enhancing the transparency of automated evaluation, laying the foundation for more interpretable and reliable AES. Our contributions can be summarized as follows:

%Interestingly, when evaluating the winning rate for rationale generation, the distilled student scorer highly matches or even surpasses the teacher LLM in relevance and accuracy aspects. Additional discussions and analysis on both scoring and rationale generation results strengthen RaDME's effectiveness in enhancing both reasoning capabilities and quality scoring. Our work는 automated evaluation의 transparency를 높이는데 기여할 첫 발자국을 내딛었다. 
% These findings highlight RaDME’s effectiveness in enhancing both reasoning capabilities and quality scoring, contributing to advancing transparency in automated evaluation. 

%RaDME delegates scoring to small expert models while training them to learn reasoning capabilities. Instead of assigning scoring tasks directly to LLMs, knowledge distillation is employed—not only due to the poor scoring performance of LLMs but also to ensure inference efficiency, resource optimization, and long-term maintainability in real-world applications.
% enhancing transparency

% This approach mitigates the poor scoring performance of LLMs while also enhancing inference and resource efficiency, and long-term maintainability in real-world applications.

\begin{itemize}
    \item We propose RaDME, a self-explainable, rationale-driven multi-trait AES that explicitly outputs reasoning with scoring, ensuring both interpretability and scoring accuracy.

    \item By providing LLMs with explicit numeric trait scores, we extract clear, coherent, and well-structured rationales, effectively supporting the distilled student model in producing high-quality explanations.
    
    \item RaDME achieves efficient and scalable AES by distilling only the reasoning capabilities of scoring-inferior LLMs, enabling a lightweight model suitable for self-explaining and scoring in real-world deployment.
    
    %We achieved efficiency in generating both rationale and scores with a smaller model fit for practices by distilling only the reasoning capacities of scoring-inferior LLMs instead of using them directly for scoring tasks.

    \item Our findings highlight the efficacy of rationale-driven scoring, revealing that scoring-first modeling notably enhances both scoring consistency and explanation quality.
    %Extensive experiments validate RaDME’s ability to generate accurate and reliable trait-wise scoring rationales while achieving competitive multi-trait scoring.
\end{itemize}

%RaDME first explicitly generates a rationale given detailed prompts including specific trait scores, followed by the corresponding trait score in a sequential manner. 
\section{Related Works}
%Highlighting the importance of explainable AES, previous studies have attempted to derive scoring outcomes by leveraging explicit grammatical or linguistic features \cite{wang2021prompt, sudoh2024automated}; however, their explainability is limited to the model’s perspective, failing to provide human-aligned explanations for understanding the assigned scores. 

%Actually, despite the recent surge in attempts to use LLMs for essay scoring, even iterative or sophisticated prompting has failed to achieve accurate scoring \cite{mizumoto2023exploring, mansour2024can, lee-etal-2024-unleashing}. 

\paragraph{LLMs for AES.}
%As LLMs continue to demonstrate exceptional performance across diverse domains, there has been a recent surge in harnessing their capabilities for automated essay scoring (AES) through zero-shot or few-shot prompting 
As LLMs continue to exhibit exceptional performance across diverse domains, their application to AES via zero-shot or few-shot prompting has gained increasing attention \cite{mizumoto2023exploring, mansour-etal-2024-large, lee-etal-2024-unleashing}. 
However, these approaches often underperform compared to fine-tuned, domain-specific models tailored for AES. ~\citet{lee-etal-2024-unleashing} used an iterative method that initially employs an LLM to generate multi-trait decomposed scoring criteria for each essay, then prompting LLM again to output textual evidence for each trait. The obtained evidence is then input for the final prompt to assign individual trait scores, which are aggregated to produce a holistic final score. Despite its tailored design and focus on holistic AES, this approach significantly underperformed relative to established AES models \cite{xie-etal-2022-automated}. Further, iterative prompting for every essay in each prompt introduced substantial computational overhead, limiting scalability and practical feasibility. Thus, instead of using LLMs directly for scoring—which proves both costly and less accurate—we propose leveraging their reasoning strengths and distilling these into a more efficient and effective scoring model.

% 따라서 우리는 Instead of directly relying on LLMs for scoring, we we employ distillation to transfer their reasoning capacity into a more efficient yet effective scoring model.  %Instead, we train the model to describe the reasoning for assessment while deriving the final score.

\begin{figure*}[t]
\centering
\includegraphics[width=0.89\textwidth]{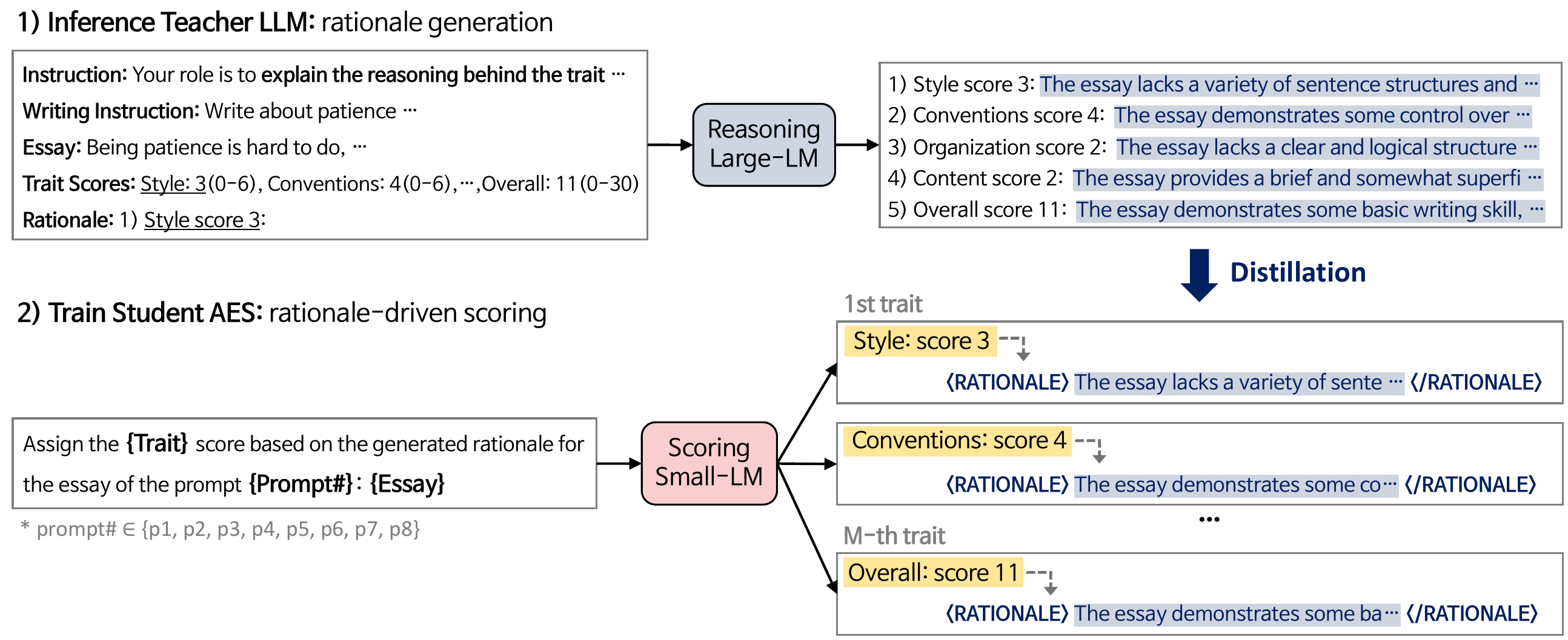}
\caption{An overview of the RaDME framework.}
\label{radme_fig}
\end{figure*}

\paragraph{Explainability in AES.}

To apply AI-based automated evaluation systems in real-world educational settings, transparency and explainability are crucial \cite{johnson2024examining}. While previous holistic AES studies have leveraged explicit grammatical or linguistic features to derive scoring outcomes \cite{wang2021prompt, sudoh2024automated}, their focus on model-based explainability often does not adequately support human understanding of the model decisions. \citet{stahl2024exploring} explored the use of LLMs for jointly providing feedback and AES, showing robust feedback quality but remarkably low scoring performance.

Recent research has shifted towards multi-trait scoring for a fine-grained evaluation of essays to provide intuitive feedback. The ArTS system \cite{do-etal-2024-autoregressive}, utilizing dependencies among traits within a text-to-text framework, has achieved robust performance in multi-trait scoring. \citet{chu2024rationale} further improved scoring accuracy by using LLM-generated rationales as an additional input for ArTS-based models. However, while their approach utilized rationales, its primary objective was to enhance AES performance rather than explainability, leaving the model unable to explicitly articulate the reasoning behind its scoring decisions. Additionally, they prompted LLMs separately for each trait, requiring iterative prompting; also, they assigned LLMs to make scoring decisions while generating rationale, which is potentially unreliable \cite{mizumoto2023exploring, lee-etal-2024-unleashing}. In contrast, we suggest yielding accurate and score-relevant rationales via score-guidance with all-trait-in-one prompting on a teacher LLM. Further, unlike their use of rationale as input and the need for LLMs in inference time, RaDME distills this capacity during training, enabling a smaller model to independently explain explicit rationales.

%Multi-trait automated essay scoring (AES) offers a fine-grained evaluation of an essay's various aspects. However, despite their success in scoring, it can not explicitly explain why specific scores are assigned to each trait, making it challenging for instructors and learners to trust the AES system outputs for practical use. Meanwhile, while large language models (LLMs) have shown competence across various generation tasks, they often underperform in AES compared to domain-specific, fine-tuned models that have learned the specific scoring patterns. Leveraging their respective strength, we propose a self-explainable rationale-driven multi-trait essay scoring (RaDME), which transfers the reasoning capabilities of LLMs to the smaller scoring-expert model, effectively teaching it to reason before scoring. In particular, we infer a teacher LLM to extract rationales for the assigned scores given a detailed prompt and then distill this reasoning into a student model. This smaller model is optimized to generate a justification first and subsequently derive a trait score based on it. Extensive experiments demonstrate that RaDME achieves both accurate and relevant reasoning while supporting high-quality multi-trait scoring, contributing to transparent AES. 
\section{RaDME}
In this work, we distinguish between using \textit{explainability} to provide feedback for behavioral changes (i.e., how to revise writing) and providing justifications for scores given the essay (i.e., explaining why the essay received a certain score). Our work focuses on the latter, aiming to enhance the transparency and trustworthiness of scoring decisions.

In human decision-making, responses are often guided by implicit rationales derived from contextual understanding. Inspired by this, we propose a self-explainable Rationale-Driven Multi-trait automated Essay scoring (RaDME) to incorporate this reasoning process by training a student model that predicts both the ground-truth trait score and its corresponding rationale, which is distilled from teacher LLM (Figure~\ref{radme_fig}).
%We incorporate rationale generation into the scoring process by training the expert scorer to self-explain reasons alongside their score decisions. 
We hypothesize that this approach encourages the model to produce scores that are more likely to be accompanied by justifications, enhancing its ability to generate a score and its well-matched explanation. RaDME is implemented within a knowledge distillation framework, leveraging LLM as the teacher and a smaller language model as the student. The student RaDME is trained to first generate a trait score and then a corresponding rationale to justify that score. It internally guides the model to assign scores grounded in reasoning, rather than generating independent explanations, effectively leveraging LLM-derived rationales while maintaining resource efficiency.

\begin{table*}
\centering
\scalebox{0.66}{%
\begin{tabular}{p{11.5cm}|p{11.5cm}}
\toprule
\rowcolor{gray!5} \textbf{LLM Prompt} & \textbf{Generation Output} \\ \midrule

Your role is to explain the reasoning behind the trait scores assigned to the \{narrative\} type essays. Please review the essay and its trait scores, and provide a one-sentence rationale for each trait score. Words beginning with ``@" have been intentionally replaced to anonymize personally identifying information in the essays.
 
\vspace{0.2cm}

\colorbox{sky!70}{Writing Instructions:} \{Write about patience. Being patient means that you are understanding and tolerant. A patient person experience difficulties without complaining. Do only one of the following: write a story about a time when you were patient OR write a story about a time when someone you know was patient OR write a story in your own way about patience.\}

\vspace{0.2cm}

\colorbox{pin!70}{Essay:} \{Being patience is hard to do, @CAPS1 I do remember a time when I was patient. This was at @CITY1. Me and my group had to wait for everyone to get back to the bus. We were the first one’s to the bus so it took a while before they came. @CAPS1 before they did I was as calm and patient as can be.\}

\vspace{0.2cm}

\colorbox{yellow!70}{Trait Scores: \{Style: 3 (0-6), Conventions: 4 (0-6), Organization: 2 (0-6),} \colorbox{yellow!70}{Content: 2 (0-6), Overall: 11 (0-30)\}}

\vspace{0.2cm}

Rationale:

\{1) Style score 3: \}

& 

Rationale:

\colorbox{yellow!70}{1) Style score 3:}  The essay lacks a variety of sentence structures and vocabulary, with the writer relying on simple sentences and basic words, which detracts from the overall style and tone of the essay.

\vspace{0.2cm}

\colorbox{yellow!70}{2) Conventions score 4:} The essay demonstrates some control over grammar, spelling, and punctuation, but there are noticeable errors, such as ``Being patience" and ``Me and my group," which affect the clarity and correctness of the writing.

\vspace{0.2cm}

\colorbox{yellow!70}{3) Organization score 2:} The essay lacks a clear and logical structure, with the writer jumping abruptly from stating the difficulty of being patient to recalling a specific experience, and failing to provide a clear conclusion or connection to the initial statement.

\vspace{0.2cm}

\colorbox{yellow!70}{4) Content score 2:} The essay provides a brief and somewhat superficial anecdote, but fails to fully develop the concept of patience, provide meaningful insights, or explore the significance of the experience, resulting in a lack of depth and substance.

\vspace{0.2cm}

\colorbox{yellow!70}{5) Overall score 11:} The essay demonstrates some basic writing skills, but is hindered by weaknesses in organization, content, and style, which limit its overall effectiveness in conveying the writer's message about patience.
\\ 
\bottomrule
\end{tabular}
}\caption[The used teacher LLM prompt and its corresponding output]{The used prompt for teacher LLM and its corresponding generation output. Notably, even without enforcing a structured format, simply indicating the first trait and its score ensured a consistent output format across all samples while effectively generating rationales for each trait score.}\label{rationale} 
\end{table*}

\definecolor{sky}{RGB}{204, 229, 255}
\definecolor{org}{RGB}{255, 229, 204}
\definecolor{gre}{RGB}{255, 255, 229}
\definecolor{pur}{RGB}{255,204,255}
\definecolor{pin}{RGB}{255,204,229}

\subsection{Extracting rationale from teacher LLMs}

To fully leverage the reasoning ability of teacher LLMs, we introduce a \textit{score-guided prompting} strategy, which explicitly informs the model of the trait-specific scores assigned to each essay. As LLMs struggle with direct numerical scoring but excel at generating explanations \cite{ryu2024guide}, we explicitly provide them with exact scores and focus them solely on generating precise rationales. Our numeric guidance strengthens the coherence between ground-truth scores and obtained rationales, ensuring alignment with actual grading patterns. This approach can effectively support the student scorer in learning to generate more relevant and well-grounded explanations.
%This process allows the extracted rationales to be high-quality and contextually relevant, which in turn helps the student model generalize better during training.

In particular, we prompt the LLM by including the following input components: {\textit{Instruction}}, which defines general roles with broad conceptual guidelines; {\colorbox{sky!70}{\textit{Writing Instruction}}}, corresponding to the essay-writing prompt that specifies the topic and theme of the learner's essay; {\colorbox{pin!70}{\textit{Essay}}($E$)}, which is the learner's written submission; and {\colorbox{yellow!70}{\textit{Trait Scores}}($S = \{s_t \mid t \in T\}$)}, comprising human-annotated trait-score (range) pairs. Given these elements, the model generates a set of rationales ($R = \{r_t \mid t \in T\}$) corresponding to the assigned trait scores. To ensure that the model generates responses in the fixed format \textit{``N) \{Trait\} score \{Score\}: \{Rationale\},"} we inform a sample format for the first trait, such as \textit{``1) Style score 3:"}. The detailed example of the used prompt and the corresponding output is described in Table~\ref{rationale}.

\subsection{Distillation for rationale-driven scoring}
% teacher LLM에서 생성된 multi-trait scores에 대한 rationale 들은 trait 단위로 분리하여 smaller scoring-expert model training 에 이용됩니다. 
%In human decision-making, answers are typically guided by implicit rationales derived from contextual understanding. 
When making decisions, humans naturally rely on implicit reasoning shaped by their understanding of the surrounding context. Motivated by this, we train the student model to predict both a ground-truth trait $t$-th score ($s_t$) and its corresponding rationale $r_t$, which is distilled from the teacher LLM. Particularly, the teacher-generated multi-trait rationales $R$ are separated by trait and employed to train a specialized scoring model optimized for multi-trait scoring with reasoning. Note that RaDME does not rely on LLMs at inference time, making it significantly more efficient and scalable for real-world deployment.

We design a unified model capable of predicting all trait scores across different prompts, leveraging an autoregressive score prediction method \cite{do-etal-2024-autoregressive}. However, distinct from their approach, which predicts all trait scores in a single sequence since it only generates trait scores, our model also produces long-form rationales. Therefore, we predict each trait independently in separate sequences to ensure stability. 

When handling essays from multiple prompts within a single model, incorporating prompt guidance in the prefix has been shown to enhance scoring accuracy, as demonstrated in ArTS \cite{do-etal-2024-autoregressive}. Building on this, since our model predicts each trait in a separate sequence, we further incorporate trait name guidance alongside prompt guidance, ensuring that the model effectively differentiates between scoring criteria, leading to more consistent and reliable predictions.

%By explicitly integrating both prompt and trait name guidance, we ensure that the model effectively distinguishes between different scoring traits while maintaining high accuracy across multiple prompts.

Building on this, since our model predicts each trait independently in separate sequences, we further incorporate trait name guidance alongside prompt guidance. To predict the $t$-th trait score, the input comprises the essay $E$, trait name $t$, and prompt number $p$, formatted as follows: \textit{``Assign the \{$t$\} score based on the generated rationale for the essay of the prompt \{$p$\}: ''}. The model then generates a {trait name} ($t$), a predicted {score} (\( \hat{s_t} \)), and a predicted {rationale} (\( \hat{r_t} \)), following the sequence:
\begin{equation}
P(t, \hat{s_t}, \hat{r_t} \mid E, t, p) = \prod_{i=1}^{N} P(y_i \mid y_{<i}, E, t, p)
\end{equation}
where $P$ indicates the probability distribution of the model's output in our autoregressive score-reasoning prediction, and $N$ is the number of tokens in the output sequence. Specifically, the model is trained to generate the output in this structured format:
\begin{equation}
{t} \ \ \  \hat{s}_{t}  \ \  \texttt{<RATIONALE>}  \ \  \hat{r}_{t}  \ \  \texttt{</RATIONALE>}
\end{equation}

%By explicitly integrating both prompt and trait name guidance, the model effectively distinguishes between different scoring traits while ensuring high accuracy and consistency across multiple prompts.

% The student model is a T5-based sequence-to-sequence model, fine-tuned to first generate a rationale  and then predict a score . Given T5’s autoregressive decoding mechanism, the order in which these components are generated significantly impacts performance. Inspired by findings that generating a rationale before the score leads to more stable predictions, we structure the decoding process as:
\section{Experiments}

\begin{table}[t]
\centering
\scalebox{0.66}{
\begin{tabular}{c|l|c|c}
\toprule
\multirow{2}{*}{\textbf{Pr}} & \multirow{2}{*}{\textbf{Traits}} & \multirow{2}{*}{\textbf{Es}} & \textbf{Score Range} \\
& & & \textbf{(Overall / Trait)} \\
\midrule
1 & Over, Cont, Org, WC, SF, Conv & 1,783 & 2 - 12 / 1 - 6 \\
2 & Over, Cont, Org, WC, SF, Conv & 1,800 & 1 - 6 / 1 - 6 \\
\midrule
3 & Over, Cont, PA, Nar, Lang & 1,726 & 0 - 3 / 0 - 3 \\
4 & Over, Cont, PA, Nar, Lang & 1,772 & 0 - 3 / 0 - 3 \\
5 & Over, Cont, PA, Nar, Lang & 1,805 & 0 - 4 / 0 - 4 \\
6 & Over, Cont, PA, Nar, Lang & 1,800 & 0 - 4 / 0 - 4 \\
\midrule
7 & Over, Cont, Org, Conv, Style & 1,569 & 0 - 30 / 0 - 6 \\
8 & Over, Cont, Org, WC, SF, Conv, Voice & 723 & 0 - 60 / 2 - 12 \\
\bottomrule
\end{tabular}}
\caption{\label{tab:dataset}
Summarized statistics of the ASAP/ASAP++ dataset. Pr: prompt number, Es: the number of essays; Over: \textit{Overall}, Cont: \textit{Content}, Org: \textit{Organization}, WC: \textit{Word Choice}, SF: \textit{Sentence Fluency}, Conv: \textit{Conventions}, PA: \textit{Prompt Adherence}, Nar: \textit{Narrativity}, Lang: \textit{Language}, Style: \textit{Style}, Voice: \textit{Voice}.}
\end{table}

\paragraph{Datasets.}

We use the most representative publicly available AES dataset, a combination of ASAP\footnote{https://www.kaggle.com/c/asap-aes} and ASAP++\footnote{https://lwsam.github.io/ASAP++/lrec2018.html} \citep{mathias2018asap++}. All comparison multi-trait scoring models are also evaluated on this dataset. ASAP++ provides human-annotated multi-trait scores for essays written in English across eight distinct prompts, offering a more granular evaluation of writing quality. Notably, ASAP++ complements the original ASAP dataset by incorporating additional trait scores that were absent in the original one. As summarized in Table~\ref{tab:dataset}, each prompt is assessed using a distinct set of writing traits with varying score ranges. While most traits appear across multiple prompts, Style and Voice are exclusively evaluated in Prompts 7 and 8, respectively, resulting in a limited number of training samples for these traits. % Our approach overcomes this limitation by employing a single unified model capable of handling all prompts and traits simultaneously.

\paragraph{Models and settings.}
For the teacher LLM, we select Llama3.1-70B \cite{dubey2024llama}, an open-source model, demonstrating competitive performance to GPT-4o \cite{hurst2024gpt} and Claude 3.5 Sonnet \cite{anthropic2024claude} on the massive multitask language understanding benchmark, to avoid the reliance on costly proprietary LLMs. As the student scoring-expert model, we employ \texttt{T5-large (770M)} \cite{2020t5}, a Transformer-based model. The generation process follows the hyperparameter settings as ArTS, using Seq2SeqTrainer with 5,000 evaluation steps, a batch size of 4, and 15 epochs. Experiments are performed on A100-SMX4-8 GPUs.

\paragraph{Evaluations and baseline models.}
In line with previous studies \cite{taghipour2016neural, do-etal-2024-autoregressive, chu2024rationale}, we perform five-fold cross-validation using the same dataset splits as their work. For evaluation, we adopt QWK, the official metric of the ASAP dataset, and report both the five-fold average scores and their standard deviations. To ensure a fair comparison, we also compute QWK scores separately for each prompt, following previous systems \cite{taghipour2016neural, do-etal-2024-autoregressive, chu2024rationale}. As baseline models, we primarily compare our approach with the robust ArTS model \cite{do-etal-2024-autoregressive} and its stronger extensions models, SaMRL \cite{do-etal-2024-autoregressive-multi} and RMTS \cite{chu2024rationale}. Details on baseline models are provided in Appendix~\ref{baseline_appendix}. % baseline 모델로 우리는 주로  robust한 \citet{do}와, 그 모델에 기반한 더 강력한 multi-trait scoring 모델들, \cite{do}, \cite{chu}랑 비교한다. 다른 구체적인 baseline은 Appendix~\ref{}에 서술하였다. 

%We primarily compared our method with the ArTS \cite{do-etal-2024-autoregressive} model, and 그 모델에 기반한 더 강력한 multi-trait scoring 모델들, \cite{do}, \cite{chu}랑 비교한다. SaMRL은 scoring-aware reinforcement learning을 도입한 모델이고, RMTS는 LLM으로부터 rationale을 뽑아서 인풋으로 사용한 모델이다. AS in existing multi-trait scorin studies, In addition, we also report the results of other multi-trait scoring models \cite{kumar2022many} and the holistic scoring models \cite{cozma-etal-2018-automated, dong2017attention} individually applied for each trait prediction. In particular, the multi-trait scoring MTL model \cite{kumar2022many} constructed each trait-specific layer and used all other trait layers auxiliary for a target trait training and prediction. The holistic scoring model, HISK, utilizes a support vector regressor paired with a histogram intersection string kernel, whereas STL-{\small LSTM} models use an LSTM-CNN-based structure; each model is iteratively deployed for independent trait scoring task. Except for the implemented ArTS*, results of other models are reported from the previous works \cite{kumar2022many, do-etal-2024-autoregressive}. 

% RMTS가 가장 좋은 모델을 가져온 것임을 말하기

\begin{table*}[t]
\centering
\scalebox{
0.71}{
\begin{tabular}{l|c|ccccccccccc|c}
\toprule
\multirow{2}{*}{\textbf{Model}} & \multirow{2}{*}{\textbf{Explainability}} & \multicolumn{11}{c|}{\textbf{Traits}} & \\
\cline{3-14}
 &&  Overall & Content & PA & Lang & Nar & Org & Conv & WC & SF & Style & Voice & AVG↑\\
\hline
HISK & \textcolor{BrickRed}{\ding{55}} & 0.718 & 0.679 & 0.697 & 0.605 & 0.659 & 0.610 & 0.527 & 0.579 & 0.553 & 0.609 & 0.489 & 0.611 \\
STL{\small-LSTM} & \textcolor{BrickRed}{\ding{55}} & 0.750 & 0.707 & 0.731 & 0.640 & 0.699 & 0.649 & 0.605 & 0.621 & 0.612 & 0.659 & 0.544 & 0.656 \\
MTL{\small-BiLSTM} & \textcolor{BrickRed}{\ding{55}} & \textbf{{0.764}} & 0.685 & 0.701 & 0.604 & 0.668 & 0.615 & 0.560 & 0.615 & 0.598 & 0.632 & {0.582} & 0.638  \\
{ArTS-large} (←) & \textcolor{BrickRed}{\ding{55}} & 0.751 & 0.730 & 0.750 & 0.701 & 0.728 & 0.675 & 0.682 & 0.680 & 0.680 & 0.715 & 0.603 & 0.700 \\
ArTS{-ind} & \textcolor{BrickRed}{\ding{55}} & 0.723 & 0.717 & \underline{0.752} & 0.695 & 0.713 & 0.649 & 0.659 & 0.662 & 0.675 & \textbf{0.722} & 0.548 & 0.683 \\
RMTS{\small-GPT} & \textcolor{BrickRed}{\ding{55}} & \underline{0.755} & \underline{0.737} & \underline{0.752} & \textbf{0.713} & \textbf{0.744} & \underline{0.682} & \underline{0.690} & \textbf{0.705} & \textbf{0.694} & 0.702 & 0.612 & \underline{0.708} \\
RMTS{\small-Llama} & \textcolor{BrickRed}{\ding{55}} & 0.754 & 0.730 & 0.749 & 0.701 & 0.737 & 0.675 & 0.684 & 0.690 & 0.684 & 0.696 & \textbf{0.640} & 0.704\\
SaMRL-large & \textcolor{BrickRed}{\ding{55}} & 0.754 & 0.735 & 0.751 & 0.703 & 0.728 & \underline{0.682} & 0.685 & 0.688 & 0.691 & 0.710 & \underline{0.627} & 0.705 \\
\midrule
{RaDME-w/o $R$} & \textcolor{BrickRed}{\ding{55}} & 0.713  & 0.700  & 0.728  & 0.655  & 0.683  & 0.636  & 0.654  & {0.647}  & 0.652  & 0.684  & {0.548}  & 0.664  \\
\textbf{{RaDME}} & \textcolor{ForestGreen}{\ding{51}} & 0.754	& \textbf{0.744} &  \textbf{0.759} & \underline{0.706} & \underline{0.736} & \textbf{0.701} & \textbf{0.692} & \underline{0.693} & \underline{0.692} & \underline{0.719} & {0.623} & \textbf{0.711} \\
\bottomrule
\end{tabular}}\caption[Trait-wise effects of RaDME averaged over prompts]{\label{tab: rad trait}
Trait-wise effects of RaDME on ASAP/ASAP++ averaged over prompts. The numerical values denote QWK scores. Bolded and underlined scores highlight the highest and the second-highest performance, respectively.   
%Five-fold averaged standard deviation is reported (\textit{SD}). ArTS* is our implemented version, and ArTS is the reported ones in \cite{do-etal-2024-autoregressive}. Higher values among the implemented baseline and ours are represented in \textbf{bold}.
}

\smallskip
\smallskip

\scalebox{
0.71}{
\begin{tabular}{l|c|cccccccc|c}
\toprule
\multirow{2}{*}{\textbf{Model}} &  \multirow{2}{*}{\textbf{Explainability}} & \multicolumn{8}{c|}{\textbf{Prompts}} & \\
\cline{3-11}
& & 1 & 2 & 3 & 4 & 5 & 6 & 7 & 8 & AVG↑ \\
\hline
HISK & \textcolor{BrickRed}{\ding{55}}& 0.674 & 0.586 & 0.651 & 0.681 & 0.693 & 0.709 & 0.641 & 0.516 & 0.644 \\
STL{\small-LSTM}  & \textcolor{BrickRed}{\ding{55}} & 0.690 & 0.622 & 0.663 & 0.729 & 0.719 & 0.753 & 0.704 & 0.592 & 0.684 \\
MTL{\small-BiLSTM}  & \textcolor{BrickRed}{\ding{55}}  & 0.670 & 0.611 & 0.647 & 0.708 & 0.704 & 0.712 & 0.684 & 0.581 & 0.665 \\
{ArTS-large} (←) & \textcolor{BrickRed}{\ding{55}} & 0.701 & 0.698 & 0.705 & \underline{0.766} & 0.725 & \underline{0.773} & \underline{0.743} & 0.635 & 0.718 \\
{ArTS-ind}  & \textcolor{BrickRed}{\ding{55}} & 0.695 & 0.679 & 0.705 & 0.762 & 0.721 & 0.756 & 0.734 & 0.578 & 0.704 \\
RMTS{\small-GPT} & \textcolor{BrickRed}{\ding{55}}  & \textbf{0.716} & {0.704} & \textbf{0.723} & \textbf{0.772} & \textbf{0.737} & 0.769 & 0.736 & 0.651 & \underline{0.726} \\
RMTS{\small-Llama}  & \textcolor{BrickRed}{\ding{55}} & 0.705 & 0.692 & 0.714 & \underline{0.766} & 0.726 & \underline{0.773} & 0.726 & \textbf{0.658} & 0.720 \\
SaMRL-large & \textcolor{BrickRed}{\ding{55}} & 0.702 & \underline{0.711} & 0.708 & \underline{0.766} & 0.722 & \underline{0.773} & \underline{0.743} & 0.649 & 0.722 \\
\midrule
{RaDME-w/o $R$}  & \textcolor{BrickRed}{\ding{55}} & 0.665  & 0.669  & 0.664  & 0.731  & 0.690  & 0.735  & 0.704  & {0.605}  & 0.683  \\
\textbf{{RaDME}} & \textcolor{ForestGreen}{\ding{51}}  & \underline{0.705} & \textbf{0.716} & \underline{0.715} & \textbf{0.772} & \underline{0.731} & \textbf{0.774} & \textbf{0.762} & \underline{0.654} & \textbf{0.729}\\
\bottomrule
\end{tabular}}
\caption[Prompt-wise effects of RaDME averaged over prompts]{\label{tab: rad prompt}
Prompt-wise effects of RaDME on ASAP/ASAP++ averaged over traits.}

\end{table*}

\section{Results}
% 놀랍게도, 우리 목표가 scoring에 대한 explainability를 향상시키는 것에 초점 맞춰졌음에도, RaDME가 기존 모델과 비교했을 때도 competitive한 성능을 보였으며, 심지어 trait-wise와 prompt-wise average 점수에선 가장 높은 QWK 성능을 보였다. 특히, 동일한 환경에서, rationale distillation을 하지 않은 것 (RaDME-w/o R) 보다 score와 rationale을 순차적으로 생성하는 것이 모든 trait과 prompt에 걸쳐서 significant 한 성능 향상을 보임으로써, RaDME의 effectiveness를 증명해준다. 
\subsection{Quality of multi-trait scoring}
Our experimental results demonstrate that RaDME achieves robust scoring performance across multiple traits and prompts while offering explainability. As shown in Table~\ref{tab: rad trait} and Table~\ref{tab: rad prompt}, RaDME outperforms other strong and state-of-the-art models, achieving the highest average QWK score in both trait-wise and prompt-wise evaluations. It is noteworthy that our method for enhancing the interpretability of scoring could also jointly improve the assessment quality. Remarkably, under the same training conditions, the sequential generation of the score and rationale (RaDME) consistently outperforms its counterpart without rationale distillation (RaDME-w/o $R$), achieving significant performance improvements across all traits and prompts. Note that RaDME-w/o $R$ is designed to isolate the effect of rationale generation, as it predicts each trait in a separate sequence as RaDME, but without generating rationales. This outcome suggests that training models to consider succeeding rationales, rather than solely relying on encoder outputs, can enhance the accuracy of score predictions, underscoring the efficacy of integrating the reasoning process into the scoring decision. 

%. This result suggests that generating rationales as an output contributes to more stable and accurate trait-specific predictions.

%Despite our primary focus on enhancing the interpretability of scoring, experimental results in Table~\ref{tab: rad trait} and~\ref{tab: rad prompt} demonstrate the strength of RaDME in multi-trait scoring. It is noteworthy that RaDME exhibits competitive performance even when compared to existing state-of-the-art models, outperforming them in both trait-wise and prompt-wise average QWK scores. Notably, under the same training conditions, the sequential generation of the score and rationale (RaDME) shows significant performance improvements across all traits and prompts compared to versions without rationale distillation (RaDME-w/o R). This outcome suggests that training models to consider succeeding rationales, rather than solely relying on encoder outputs, can enhance the accuracy of score predictions, underscoring the efficacy of integrating the reasoning process into the scoring decision. 

% 이 결과는 모델이 순수하게 encoder output만 보고 점수를 결정하도록 학습되는 것보다, 뒤따라오는 이유를 고려하도록 학습되는 것이 점수 예측을 assist 할 수 있음을 암시해준다  
% 모델이 점수를 예측할 때, 뒤따라오는 "이유"까지 고려해야 하므로, 점수 예측 과정에서 미래의 정보(future token)를 간접적으로 학습할 가능성이 있음. 이 과정에서, 모델이 점수를 예측할 때 이후에 설명할 수 있는 더 합리적인 점수를 선택하도록 학습될 가능성이 있음.

\paragraph{Single vs. sequential trait prediction.}
As observed in the comparison between ArTS (i.e., predicting all traits in a sequence) and ArTS-ind (i.e., using individual models for separated trait sequence), incorporating trait-wise dependencies within an autoregressive decoding strategy has been revealed to improve performance \cite{do-etal-2024-autoregressive}. However, in our experiments on a single fold, predicting the score-rationale sequence for all traits within a single forward pass resulted in significantly lower performance, with a trait-wise average of 0.454 and a prompt-wise average of 0.504. Since RaDME generates a rationale for each trait score, predicting all traits at once can cause instability in subsequent trait predictions. Additionally, as each essay can be evaluated on up to seven traits, later predictions may suffer from information loss regarding the essay content. This comparison emphasizes that in scenarios requiring explanations, our approach—predicting one trait at a time—can better assist the model in understanding the contexts to make precise decisions. It is noteworthy that despite ArTS, SaMRL \cite{do-etal-2024-autoregressive-multi}, and RMTS \cite{chu2024rationale} being explicitly designed to leverage trait dependencies, giving them an inherent advantage in multi-trait AES tasks, RaDME still outperforms them. This result provides strong evidence that rationale distillation itself can enhance the model’s decision-making capabilities, further validating our approach.

%ArTS와 ArTS-ind를 비교했을 때 알 수 있듯이 autroregressive decoding strategy 내에서 triat간의 dependcy를 반영해주는게 성능 면에서는 더 효과적인 것으로 알려져있다 \cite{do}. 하지만, 한 fold에서 전체 trait에 대해 score-rationale sequence를 한번에 예측하도록 했을 때, trait-wise 평균 0.454, prompt-wise 평균 0.504 정도로 훨씬 낮은 성능을 보였다. RaDME의 경우, 각 trait 점수마다 rationale을 생성하기 때문에, 모든 trait을 한 번에 예측하면 이어지는 trait 예측에서 오히려 불안정성이 증가할 수 있다. 또한 한 에세이당 최대로 7개의 trait에 대해 평가될 수 있기 때문에 뒤로 갈수록 에세이 정보에 대한 손실이 일어날 수 있다. 이 비교 결과는 explain을 해야하는 상황에서는 한번에 하나의 trait을 판단하도록 하는 우리의 접근이 더 효과적임을 보여준다. 또한, 여기서 주목할만한 점은, ArTS, SaMRL, RMTS 모두 trait-dependency를 활용하도록 학습되어서 AES task 측면에서는 우리 모델보다 훨씬 유리함에도, RaDME가 더 높은 성능을 보였다는 점이다. 이는 rationale distillaiton 자체가 모델의 판단력에 도움을 준다는 것을 증명해준다. 

\begin{figure*}[t]
\centering
\includegraphics[width=0.8\linewidth]{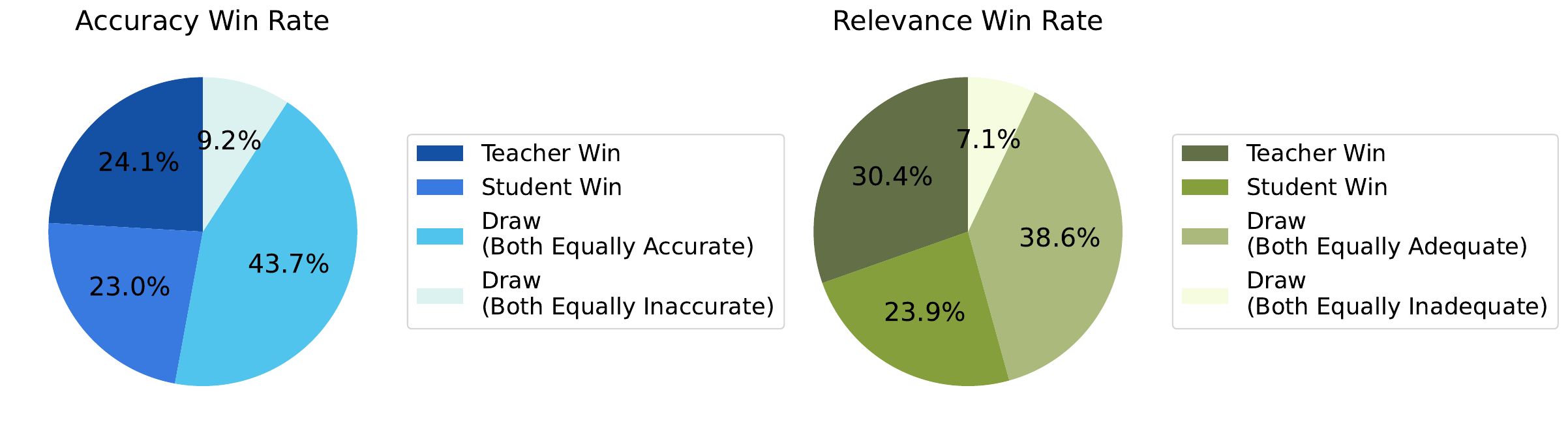}
\caption[Evaluation of win rates for accuracy between generated rationales]{Evaluation of win rates for accuracy and relevance between rationales generated by the student model and those generated by the LLM on the test set.}
\label{fig: vote}
\end{figure*}

\paragraph{Self-explaining vs. injecting rationales.}
% RMTS는 trait-wise LLM agent를 따로 두고, LLM에게 에세이에 대한 점수와 그 근거를 생성하도록 prompting을 하고, 생성된 근거를 에세이만 인풋으로 넣던 기존 ArTS 모델에, 추가로 넣어주는 방법론이다. rationale을 생성한다는 점에서 공통점이 있지만, 1) RaDME는 오로지 essay 인풋만 받고, scoring model이 직접 rationale을 생성하도록 teaching한다는 점 2) 그렇기 때문에 우리는 inference때 LLM 없이 small student 모델만 있으면 된다는 점에서 큰 차이가 있다. 이런면에서 우리 모델이 실제 상황에서 널리 사용될 수 있는 효율성과 확장성을 갖추었다면, scoring performance 입장에선 우리의 self-explaining 방식과 RMTS의 raionale-injection 방식이 어떻게 효과적일지를 explore 해보았다. 즉, AES 성능엔 output으로서의 rationale이 좋을까 input으로서의 rationale이 좋을까를 살펴 보았다. 결과적으로,  

We investigate the impact of rationales as input (i.e., RMTS \cite{chu2024rationale}), versus generating rationales as an output (RaDME). While RMTS incorporates rationales as additional context for AES, our RaDME generates them internally, allowing the model to self-explain its scoring decisions. Despite RMTS based on trait dependencies and rationale injection, which highly benefits scoring, RaDME’s self-explaining mechanism surpasses it in overall scoring performance. The results suggest that training the model to learn reasoning alongside scoring not only enhances model interpretability but also improves scoring efficacy. Notably, our system achieves higher performance in \textit{Content}, \textit{Prompt Adherence}, and \textit{Organization}—traits that require a comprehensive understanding of the essay's contextual coherence and logic rather than just identifying isolated elements. Beyond its robustness in scoring, a key advantage of RaDME is that it does not require LLMs at inference time, making it significantly more efficient and scalable for practices.

\subsection{Quality of the generated rationales}

To validate the quality of the rationale generated by the RaDME, we measured the winning rate using GPT-4o, randomly selecting 1,000 samples. The evaluation involved comparing two rationales: the teacher LLM's rationale and the RaDME-generated (i.e., student model's) rationale. Evaluators selected one of four possible outcomes: Teacher—Rationale 1—Win, Student—Rationale 2—Win, Draw (Both Good), or Draw (Both Poor). A detailed example of the prompt is provided in Appendix~\ref{appendix_prompts} (Table~\ref{winning_rate}). We evaluated two dimensions: \textit{accuracy}, which measures whether the rationale contains only correct information, and \textit{relevance}, which assesses whether the rationale adequately includes the necessary information. 

Figure~\ref{fig: vote} results showed that only 9.2\% of samples were rated as poor for both rationales. Surprisingly, 66.7\% of the student rationales have results more accurate or equally accurate compared to the teacher-generated ones, suggesting that the scoring procedure itself may influence the quality of the rationale. For relevance, only 8.8\% of the samples were rated as poor for both models, and the majority (62.5\%) were judged to be better or as good as the teacher model. These findings demonstrate that RaDME, even with an efficient student model, can generate effective rationales that accurately convey the reasoning behind essay scores, making it suitable for practical settings. Detailed qualitative analyses for rationales are provided in Appendix~\ref{qualitative}.

\begin{figure}
\centering
\includegraphics[width=0.93\linewidth]{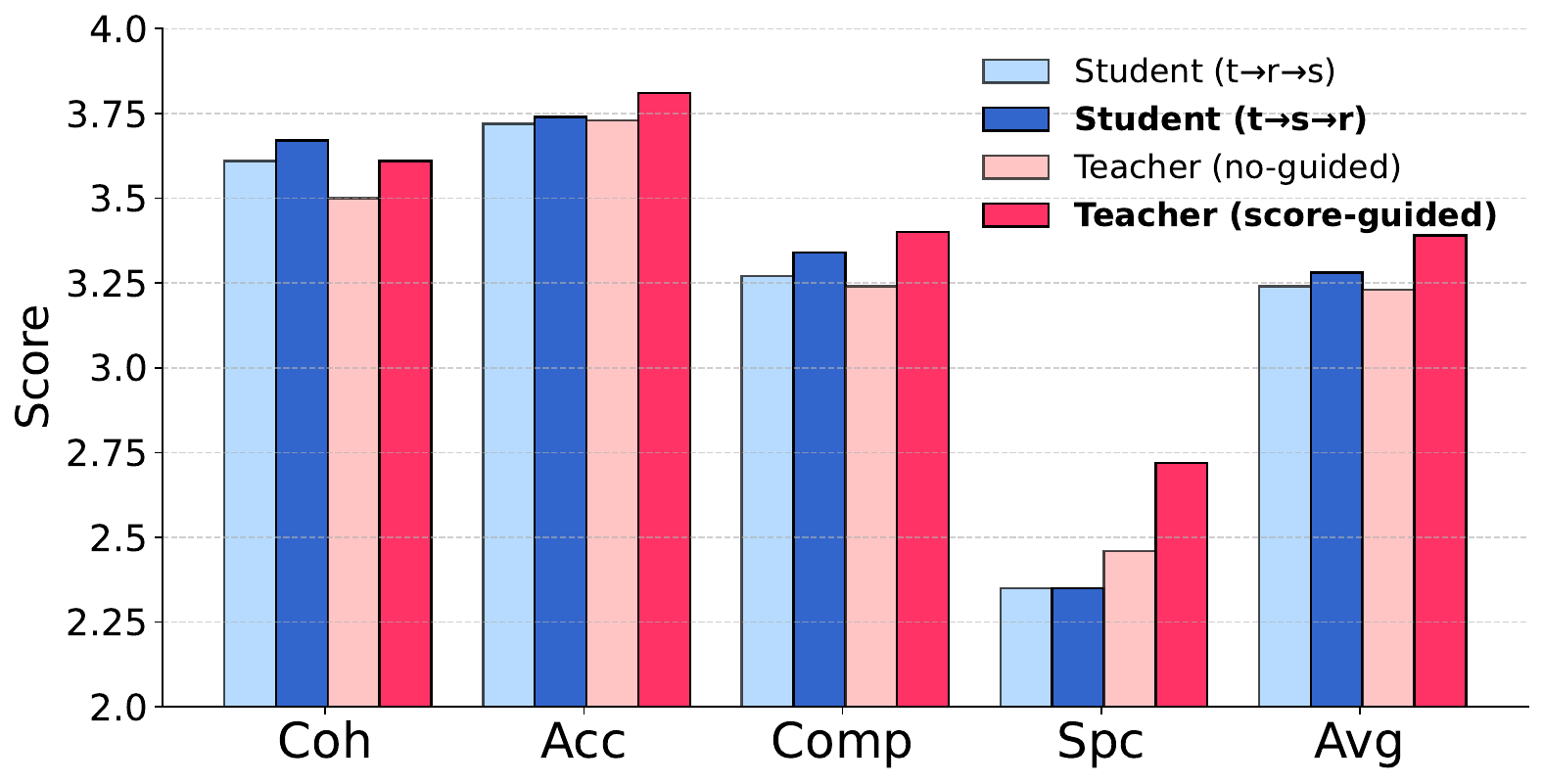}
\caption{Evaluation results with G-Eval.}%: \textit{Coherence}, \textit{Accuracy}, \textit{Completeness}, and \textit{Specificity}.}
\label{fig: geval}
\end{figure}

\section{Discussions}

\begin{table*}[t]
\centering
\scalebox{
0.70}{
\begin{tabular}{l|ccccccccccc|c}
\toprule
\multirow{2}{*}{\textbf{Model}} & \multicolumn{11}{c|}{\textbf{Traits}} & \\
\cline{2-13}
 &  Overall & Content & PA & Lang & Nar & Org & Conv & WC & SF & Style & Voice & AVG↑\\
\hline
Teacher (No-guided) & 0.405 & 0.406 & 0.358 & 0.381 & 0.353 & 0.471  & 0.395 & 0.459  & 0.456 & 0.235 & 0.491 & 0.401 \\
\midrule

{RaDME} ($t\rightarrow r\rightarrow s$) & 0.728  & 0.727 & 0.750  & 0.671  & 0.720  & 0.678 & 0.686 & 0.642  & 0.673  & 0.702  & 0.524 & 0.682 \\
\rowcolor{sky!47} \textbf{{RaDME} ($t\rightarrow s\rightarrow r$)} &\textbf{0.754}	& \textbf{0.744} &  \textbf{0.759} & \textbf{0.706} & \textbf{0.736} & \textbf{0.701} & \textbf{0.692} & \textbf{0.693} & \textbf{0.692} & \textbf{0.719} & \textbf{{0.623}} & \textbf{0.711} \\
%\rowcolor{pink!33} \textbf{Teacher (Score-guided) }& 0.996	& 0.999	& 1.000	& 1.000	& 1.000	& 1.000	& 1.000	& 1.000	& 1.000	& 1.000	& 1.000 & 1.000 \\
\bottomrule
\end{tabular}}\caption[Trait-wise effects of RaDME averaged over prompts]{\label{tab: rad ablation}
Trait-wise QWK results of Teacher w/o score-guidance, RaDME ($t\rightarrow r\rightarrow s$), and the original RaDME ($t\rightarrow s\rightarrow r$). Bolded scores highlight the highest performance. 
}
\smallskip
\smallskip
\scalebox{
0.70}{
\begin{tabular}{l|cccccccc|c}
\toprule
\multirow{2}{*}{\textbf{Model}} & \multicolumn{8}{c|}{\textbf{Prompts}} & \\
\cline{2-10}
 & 1 & 2 & 3 & 4 & 5 & 6 & 7 & 8 & AVG↑ \\
\hline
 Teacher (No-guided) & 0.407 & 0.526 & 0.353 & 0.412 & 0.395 & 0.353 & 0.331 & 0.425 & 0.400 \\
  \midrule
 {RaDME} ($t\rightarrow r\rightarrow s$) & 0.699 & 0.688 & 0.696 & 0.763 & 0.708 & 0.747 & 0.740 & 0.605 & 0.706 \\
\rowcolor{sky!47} \textbf{{RaDME} ($t\rightarrow s\rightarrow r$)}  & \textbf{0.705} & \textbf{0.716} & \textbf{0.715} & \textbf{0.772} & \textbf{0.731} & \textbf{0.774} & \textbf{0.762} & \textbf{0.654} & \textbf{0.729}\\
%\rowcolor{pink!33} \textbf{Teacher (Score-guided)} & \textbf{1.000} & \textbf{1.000} & \textbf{1.000} & \textbf{0.999 }& \textbf{1.000} & \textbf{0.999} & \textbf{1.000} & \textbf{0.995} & \textbf{0.999} \\
\bottomrule
\end{tabular}}
\caption[Prompt-wise effects of RaDME averaged over prompts]{\label{tab: rad ablation prompt}
Prompt-wise QWK results of Teacher w/o score-guidance, RaDME ($t\rightarrow r\rightarrow s$), and the original RaDME.}

\end{table*}

\definecolor{gree}{RGB}{102,204,0}

\paragraph{Effect of score-rationale generation order.}\label{sec: geval}
In our framework, RaDME generates a trait score first, followed by its rationale. To examine whether this prediction order is optimal, we conduct additional experiments where the model first generates the rationale and then the score.

Our results in Table~\ref{tab: rad ablation} and Table~\ref{tab: rad ablation prompt} demonstrate that the score-first approach consistently outperforms the rationale-first approach across both trait-wise and prompt-wise evaluations. Determining the score before generating the rationale anchors the explained output to a concrete decision, ensuring alignment between the two and leading to stable predictions. Contrarily, without a predefined score to guide the rationale, the model may struggle to produce explanations that align with appropriate numerical assessment, resulting in greater variance in score predictions. These results suggest that in AES tasks where both accuracy and explainability are vital, deciding the score first is more effective.

% RaDME는 평가할 측면에 대해 점수를 먼저 생성하고 그 후 rationale을 생성하는데, 이 방식이 적합한지 examine 하기 위해 우리는 추가로 rationale을 먼저 생성하고 점수를 생성하는 방법으로도 실험을 진행해보았다.    

We further investigated whether rationale quality itself benefits from being predicted first (Figure~\ref{fig: geval}; \colorbox{sky!80}{blue}). For evaluation, we utilized G-Eval \cite{liu-etal-2023-g}, an evaluation framework proven robustness in natural language generation using GPT-4. We compared RaDME ($t\rightarrow s\rightarrow r$) with its reverse variant ($t\rightarrow r\rightarrow s$) to assess the impact of generation order on rationale quality. G-Eval automatically generates a Chain-of-Thought (CoT) evaluation based on the target dimension’s criteria. We define four key dimensions for assessing rationale quality: \textit{Coherence}, \textit{Accuracy}, \textit{Completeness}, and \textit{Specificity}. \textit{Coherence} evaluates whether the rationale presents a clear, structured, and logically connected explanation, with higher scores indicating smoother and well-organized reasoning. \textit{Accuracy} measures how correctly the rationale justifies the assigned score. \textit{Specificity} assesses the level of detail, where higher scores reflect more precise explanations. \textit{Completeness} examines whether the rationale fully addresses all relevant aspects of the essay trait being scored. Following these criteria, we generate a CoT-based assessment and assign scores on a 1–5 scale for randomly selected 100 samples. Specific evaluation methods and the prompts used are detailed in Appendix~\ref{sec: appendix-geval_prompt}. Interestingly, comparison results (Figure~\ref{fig: geval}) suggest predicting the score first also leads to superior rationale quality across all dimensions. The results indicate that the preceding score decision can induce more structured and coherent reasoning, while the opposite increases variance in explanation quality. %For rationale evaluation using G-Eval, we randomly selected 100 samples.

\paragraph{Impact of score-guided prompting.}
To fully leverage the capabilities of the teacher LLM, we proposed a \textit{score-guided prompting} strategy. To verify whether our guidance effectively led to high-quality outputs, we conducted a comparative analysis with no guidance prompting. As shown in Table~\ref{tab: rad ablation} and Table~\ref{tab: rad ablation prompt}, the teacher model without score guidance (Teacher No-Guided), where the LLM is directly tasked with generating both scores and rationales, exhibits poor scoring performance. This result aligns with previous research findings \cite{lee-etal-2024-unleashing}, which highlight the inherent limitations of LLMs in accurate score prediction, supporting the necessity of our score-guidance method.

We also conducted further evaluations of the generated rationales using G-Eval \cite{liu-etal-2023-g}, following the same four evaluation criteria defined earlier. Our results (Figure~\ref{fig: geval}; \colorbox{red!20}{red}) reveal that providing exact trait scores significantly enhances rationale quality across all evaluation dimensions. Specifically, our score-guided generation yields rationales that are more coherent, accurate, and complete compared to those generated without explicit score guidance. Notably, we observe a substantial improvement in \textit{Specificity}, suggesting that grounding rationale generation in predefined correct answers enhances the model’s ability to produce more precise and well-structured explanations. These findings highlight the critical role of explicit score guidance in improving rationale generation, which then subsequently affects the distillation efficacy for optimizing the student model.

% 우리는 Teacher LLM의 능력을 최대한 끌어내기 위해 score-guidance strategy를 제안했었다. 실제로 우리의 guidance가 high quality의 output를 이끌어냈는지를 검증하기 위해, Section 6.1에서와 마찬가지로 G-Eval \cite{liu-etal-2023-g}을 기반으로 동일한 4가지의 criteria 에 대해 비교 평가를 진행했다 (Figure~\ref{}). 결과적으로, 모든 aspects에서 exact trait scores들을 넘겨주는 것이 그렇지 않은 것보다 훨씬 coherence하고 accurate하면서 completness한 rationale을 생성해 내었다. 무엇보다도, simplicity에서 큰 폭으로 좋은 성능을 보였는데, 이는 실제 매겨진 정답에 기반해서 rationale을 생성하도록 하는 것이 구체적인 답변을 생성하는데 도움이 됨을 시사한다.       

\section{Conclusion}

We propose RaDME, a self-explainable, rationale-driven multi-trait AES method that enhances both transparency and scoring accuracy. Unlike existing AES systems that lack an explanation for assigned scores, RaDME explicitly generates rationales alongside trait scores, making its decisions interpretable. By distilling LLMs' reasoning capabilities into a scoring-efficient student model, RaDME achieves both high-quality scoring and clear, coherent explanations. Our extensive experiments reveal that while LLMs struggle with direct scoring, they excel in rationale generation when provided with precise numerical scores. RaDME successfully integrates this strength, producing accurate, well-structured, and detailed rationales while maintaining outstanding scoring performance.

\section*{Limitations}
While RaDME demonstrates strong technical and empirical performance, its practical impact in real-world educational environments, particularly in human-centered interactions, remains an area for future exploration. Assessing its effect on students in interactive settings would further enhance the significance of this research. Nevertheless, we believe this limitation is not unique to RaDME but applies to most recent AES systems, which often lack explicit evaluation of human-centered effects. Therefore, future research could extensively explore how educators and students perceive, interpret, and utilize rationale-based feedback provided by RaDME, presenting an important direction for further investigation.

 %While RaDME demonstrates strong technical and empirical performance, its practical impact in real-world educational environments, particularly regarding to human-centered-interaction, remains as a future work. Assessing its practical effects on students in interactive settings would further strengthen the significance of this research. Nevertheless, we believe this limitation is not unique to RaDME but applies to most recent AES systems, which often lack explicit evaluation of human-centered effects. Thus, future consideration of how educators and students perceive, interpret, and utilize rationale-based feedback provided by RaDME, might be an interesting research direction.

\section*{Ethical Statement}
We used publicly available datasets of automated essay scoring in this study, ensuring compliance with ethical guidelines and data usage policies. The dataset does not contain personally identifiable or sensitive information.

\section*{Acknowledgments}
This work was supported by the IITP (Institute of Information \& Coummunications Technology Planning \& Evaluation)-ITRC (Information Technology Research Center) grant funded by the Korea government (Ministry of Science and ICT) (IITP-2025-RS-2024-00437866) and by Institute of Information \& communications Technology Planning \& Evaluation (IITP) grant funded by the Korea government(MSIT) (No.RS-2019-II191906, Artificial Intelligence Graduate School Program (POSTECH)).

\bibliography{main}

\appendix

\section{Baseline Models} \label{baseline_appendix}

We primarily compare our method with ArTS \cite{do-etal-2024-autoregressive} and its enhanced multi-trait scoring extensions, SaMRL \cite{do-etal-2024-autoregressive-multi}, and RMTS \cite{chu2024rationale}. Among them, SaMRL incorporates scoring-aware reinforcement learning, while RMTS leverages rationales extracted from LLMs as additional input.

Following prior multi-trait scoring studies, we also report results for other multi-trait scoring models \cite{kumar2022many} and holistic scoring models \cite{cozma-etal-2018-automated, dong2017attention}, where each holistic model is applied independently for trait-specific predictions. Specifically, the multi-trait scoring MTL model \cite{kumar2022many} employs trait-specific layers while using all other trait layers as auxiliary components for target trait training and prediction. For holistic scoring, HISK utilizes a support vector regressor with a histogram intersection string kernel, while STL-{\small LSTM} applies an LSTM-CNN-based structure, with each model iteratively deployed for independent trait scoring.

\section{Qualitative Analysis of Rationales}\label{qualitative}

\begin{figure*}
\centering
\includegraphics[width=0.95\linewidth]{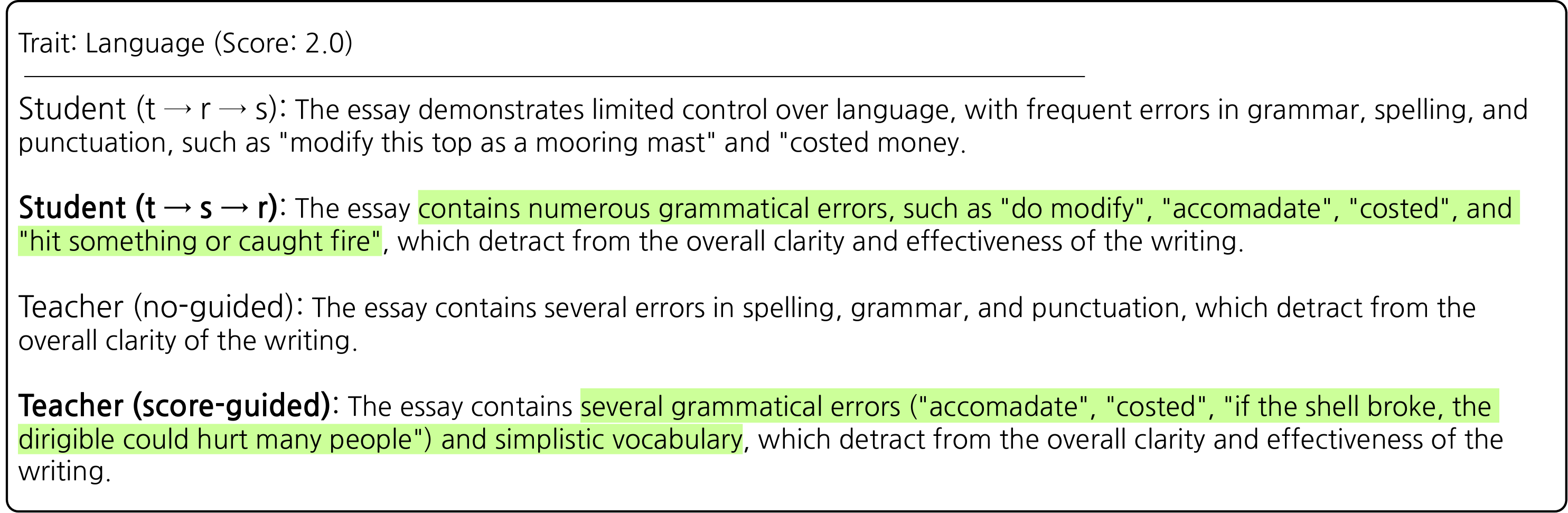}
\caption{Comparison of rationales generated by different models for the \textit{Language} trait, in the case of a score of 2. Bolded models represent our proposed methods, while green highlights indicate well-specified phrases within the rationales.}
\label{fig: bad1}
\end{figure*}

\begin{figure*}
\centering
\includegraphics[width=0.95\linewidth]{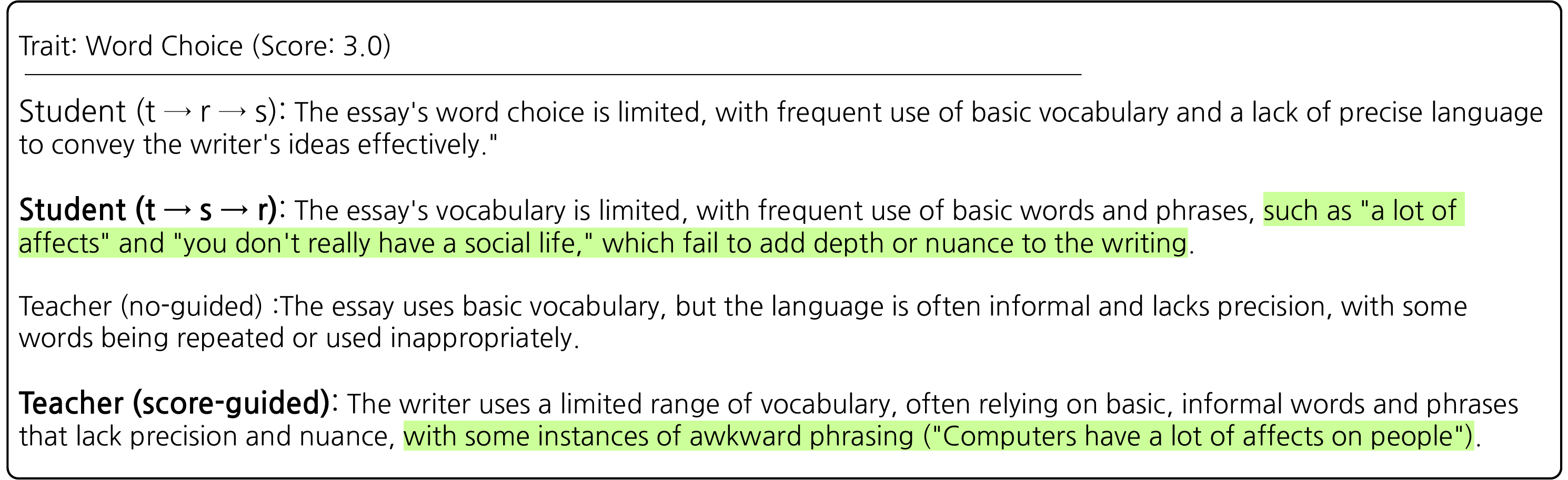}
\caption{Comparison of rationales generated by different models for the \textit{Word Choice} trait, in the case of a score of 3.
}
\label{fig: bad2}
\end{figure*}

\begin{figure*}
\centering
\includegraphics[width=0.95\linewidth]{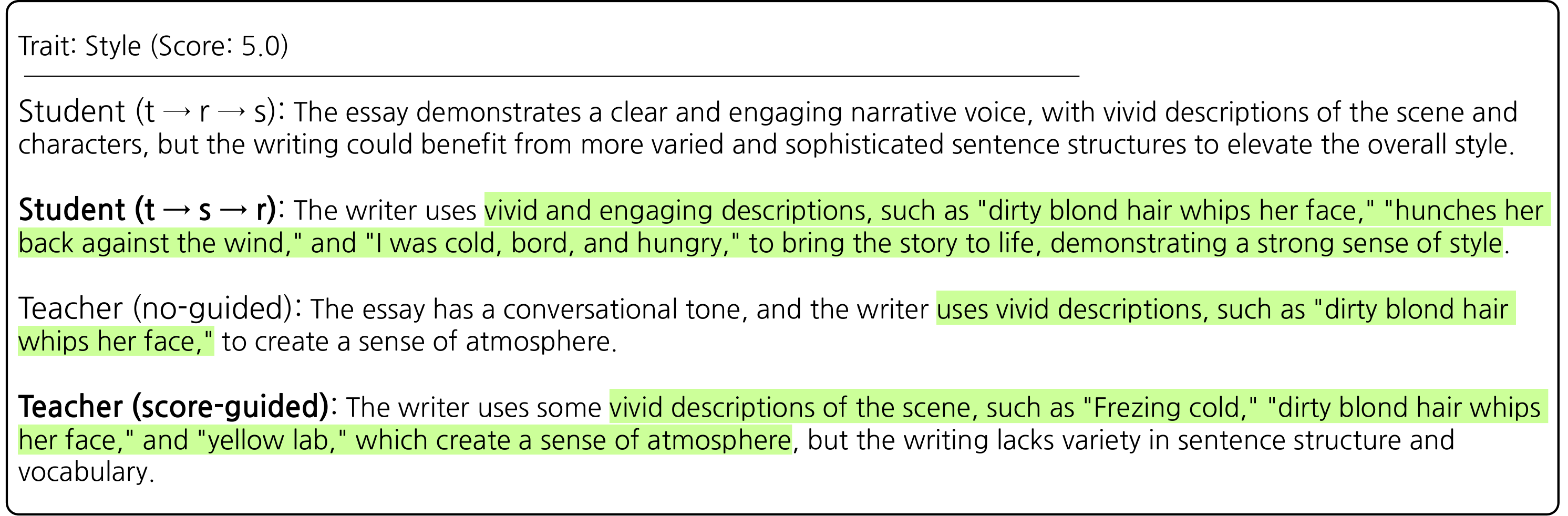}
\caption{Comparison of rationales generated by different models for the \textit{Style} trait, in the case of a score of 5.}
\label{fig: good1}
\end{figure*}

Figure~\ref{fig: bad1},~\ref{fig: bad2}, and ~\ref{fig: good1} compare the actual generated rationales from four different models, discussed in Section 6. As illustrated in these figures, our proposed methods yield more detailed and clearly articulated rationales.

\section{Detailed Prompts} \label{appendix_prompts}

\subsection{Prompt for the winning rate evaluation}
The prompt used to evaluate the winning rate between the teacher’s and the student’s rationales is described in Table~\ref{winning_rate}. The results are shown in Figure~\ref{fig: vote} and discussed in Section 5.2.

\begin{table*}
\centering
\scalebox{0.72}{%
\begin{tabular}{p{10.2cm} | p{10.2cm}}
\toprule
\rowcolor{gray!5} \textbf{Prompt for Accuracy} & \textbf{Prompt for Relevance}\\ \midrule

[system prompt] 

Your task is to evaluate which rationale most \textbf{accurately} explains the assigned scores for the essay.

[input prompt]

Please review the essay, trait score, and each rationale carefully, and then choose one of the following options:  

\vspace{0.2cm}

\#\#\# Writing Instruction: \{instruction\}

\#\#\# Essay: \{essay\}

\#\#\# \{trait\} Trait Score: \{score\}

\#\#\# Rationale 1: \{rationale1\}

\#\#\# Rationale 2: \{rationale2\}

1. Rationale 1 most \textbf{accurately} explains the trait quality of the essay. 

2. Rationale 2 most \textbf{accurately} explains the trait quality of the essay. 

3. Draw (both rationales are equally \textbf{accurate}): Both rationales provide an equally \textbf{accurate} explanation of the assigned scores. 

4. Draw (both rationales are equally \textbf{inaccurate}): Both rationales fail to provide an \textbf{accurate} explanation of the assigned scores. 

Provide only the corresponding option number: 

\vspace{0.1cm} &

[system prompt] 

Your task is to evaluate which rationale most \textbf{adequately} explains the assigned scores for the essay.

[input prompt]

Please review the essay, trait score, and each rationale carefully, and then choose one of the following options:  

\vspace{0.2cm}

\#\#\# Writing Instruction: \{instruction\}

\#\#\# Essay: \{essay\}

\#\#\# \{trait\} Trait Score: \{score\}

\#\#\# Rationale 1: \{rationale1\}

\#\#\# Rationale 2: \{rationale2\}

1. Rationale 1 most \textbf{adequately} explains the trait quality of the essay. 

2. Rationale 2 most \textbf{adequately} explains the trait quality of the essay. 

3. Draw (both rationales are equally {\textbf{adequate}}): Both rationales provide an equally \textbf{adequate} explanation of the assigned scores. 

4. Draw (both rationales are equally \textbf{inadequate}): Both rationales fail to provide an \textbf{adequate} explanation of the assigned scores. 

Provide only the corresponding option number: 

\vspace{0.1cm}
\\ 
\bottomrule
\end{tabular}}
\caption{Evaluation prompts for comparing the winning rates of rationales generated by teacher and student models. An example of the \textit{accuracy} aspect. 
 }\label{winning_rate} 
\end{table*}

\subsection{Prompt for the teacher LLM without score-guided prompting}

The prompt used to generate the rationale without our score-guided prompting strategy, as a comparison results, are shown in Table~\ref{fake_rationale}. The results are shown in Table~\ref{tab: rad ablation} and Table~\ref{tab: rad ablation prompt} and discussed in Section 6.

\begin{table*}[t]
\centering
\scalebox{0.67}{%
\begin{tabular}{p{11cm}|p{11cm}}
\toprule
\rowcolor{gray!5} \textbf{LLM Prompt} & \textbf{Generation Output} \\ \midrule
\vspace{0.01cm}
Your role is to assign trait scores to the narrative type essay and explain the reasoning behind the assigned scores. Please read the essay and assign its trait scores, each followed by a one-sentence rationale. Words beginning with '@' have been intentionally replaced to anonymize personally identifying information in the essays.

\vspace{0.2cm}

\colorbox{sky!70}{Writing Instructions:} \{Write about patience. Being patient means that you are understanding and tolerant. A patient person experience difficulties without complaining. Do only one of the following: write a story about a time when you were patient OR write a story about a time when someone you know was patient OR write a story in your own way about patience.\}

\vspace{0.2cm}

\colorbox{pin!70}{Essay:} \{Being patience is hard to do, @CAPS1 I do remember a time when I was patient. This was at @CITY1. Me and my group had to wait for everyone to get back to the bus. We were the first one’s to the bus so it took a while before they came. @CAPS1 before they did I was as calm and patient as can be.\}

\vspace{0.2cm}

\colorbox{yellow!70}{Traits (Score ranges): \{Style (0-6), Conventions (0-6), Organization} \colorbox{yellow!70}{(0-6), Content (0-6), Overall (0-30)\}}

\vspace{0.2cm}

Please score each trait only once with the following format: `trait score [score]: [rationale].'

\vspace{0.2cm}

Assigned Scores and Rationales:

\{1) Style score \}

\vspace{0.1cm} & 

\vspace{0.01cm}

Assigned Scores and Rationales: 

\colorbox{yellow!70}{1) Style score 2:} The writing lacks a clear and engaging tone, and the language is simple and repetitive.

\vspace{0.2cm}

\colorbox{yellow!70}{2) Conventions score 3:} The essay contains several grammatical errors, including incorrect verb tense and subject-verb agreement, but the writer demonstrates some control over basic sentence structure.

\vspace{0.2cm}

\colorbox{yellow!70}{3) Organization score 3:} The narrative is loosely structured and lacks transitions between sentences, but the writer attempts to tell a story with a clear beginning and end.

\vspace{0.2cm}

\colorbox{yellow!70}{4) Content score 3:} The essay provides a brief anecdote about being patient, but the story is thin and lacks descriptive details that would make it more engaging and memorable.

\vspace{0.2cm}

\colorbox{yellow!70}{5) Overall score 11:} The essay demonstrates some effort to tell a story about patience, but it is marred by errors in grammar, syntax, and organization, and lacks a clear and engaging tone.

\vspace{0.1cm}
\\ 
\bottomrule
\end{tabular}
}\caption[The used teacher LLM prompt and its corresponding output]{The prompt used for the teacher LLM without our score-guidance and its corresponding generation output. 
Although the ground-truth scores are \textit{Style: 3, Conventions: 4, Organization: 2, Content: 2, Overall: 11}, the scores generated without score guidance deviate significantly from the actual values. This result contrasts with the findings presented in Table~\ref{rationale}.} \label{fake_rationale} 
\end{table*}

\subsection{Prompts for G-Eval}
\label{sec: appendix-geval_prompt}

\begin{figure*}
\centering
\includegraphics[width=0.95\linewidth]{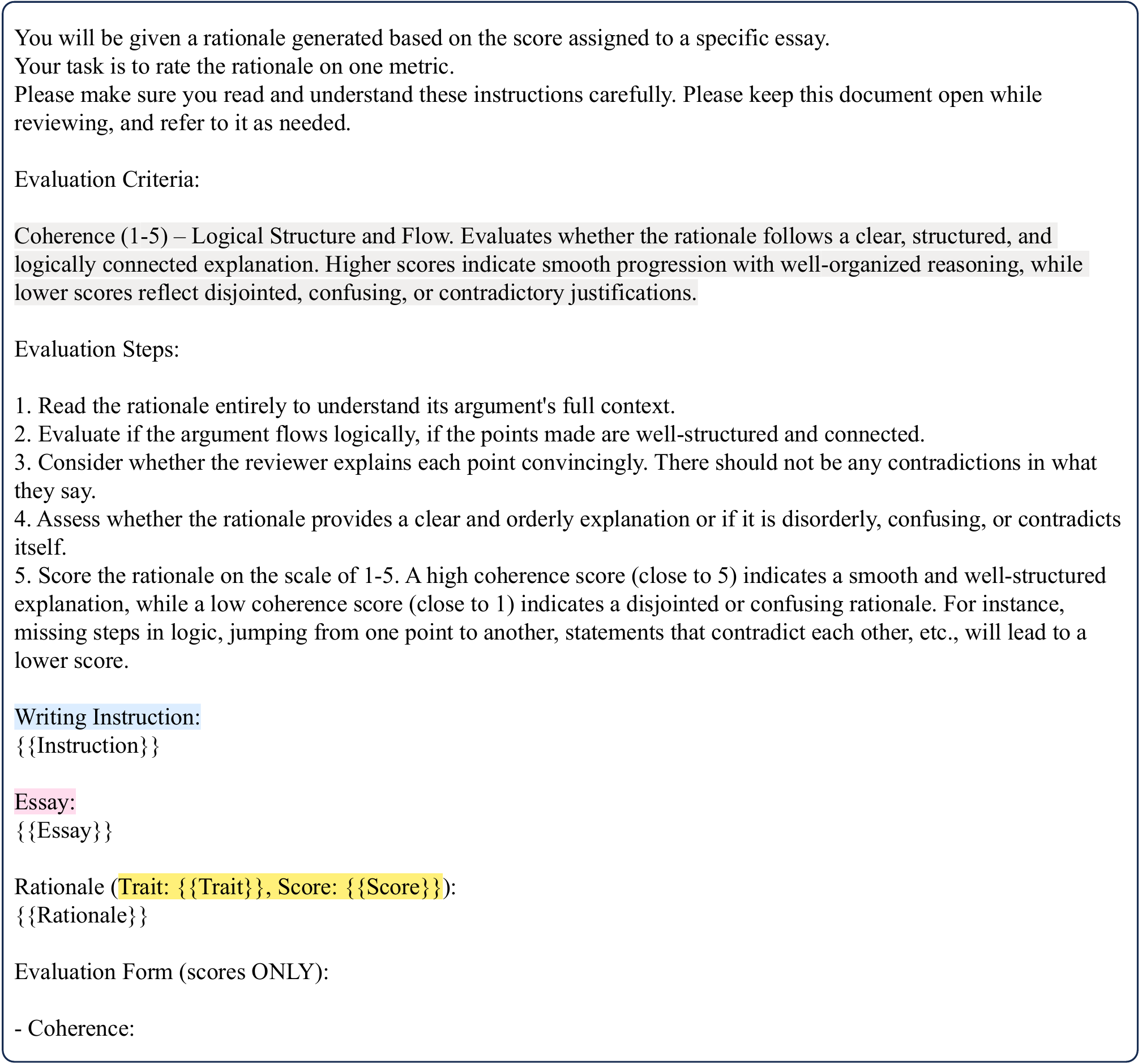}
\caption{Evaluation for \textit{Coherence}.}
\label{fig: geval-coherence}
\end{figure*}

\begin{figure*}
\centering
\includegraphics[width=0.95\linewidth]{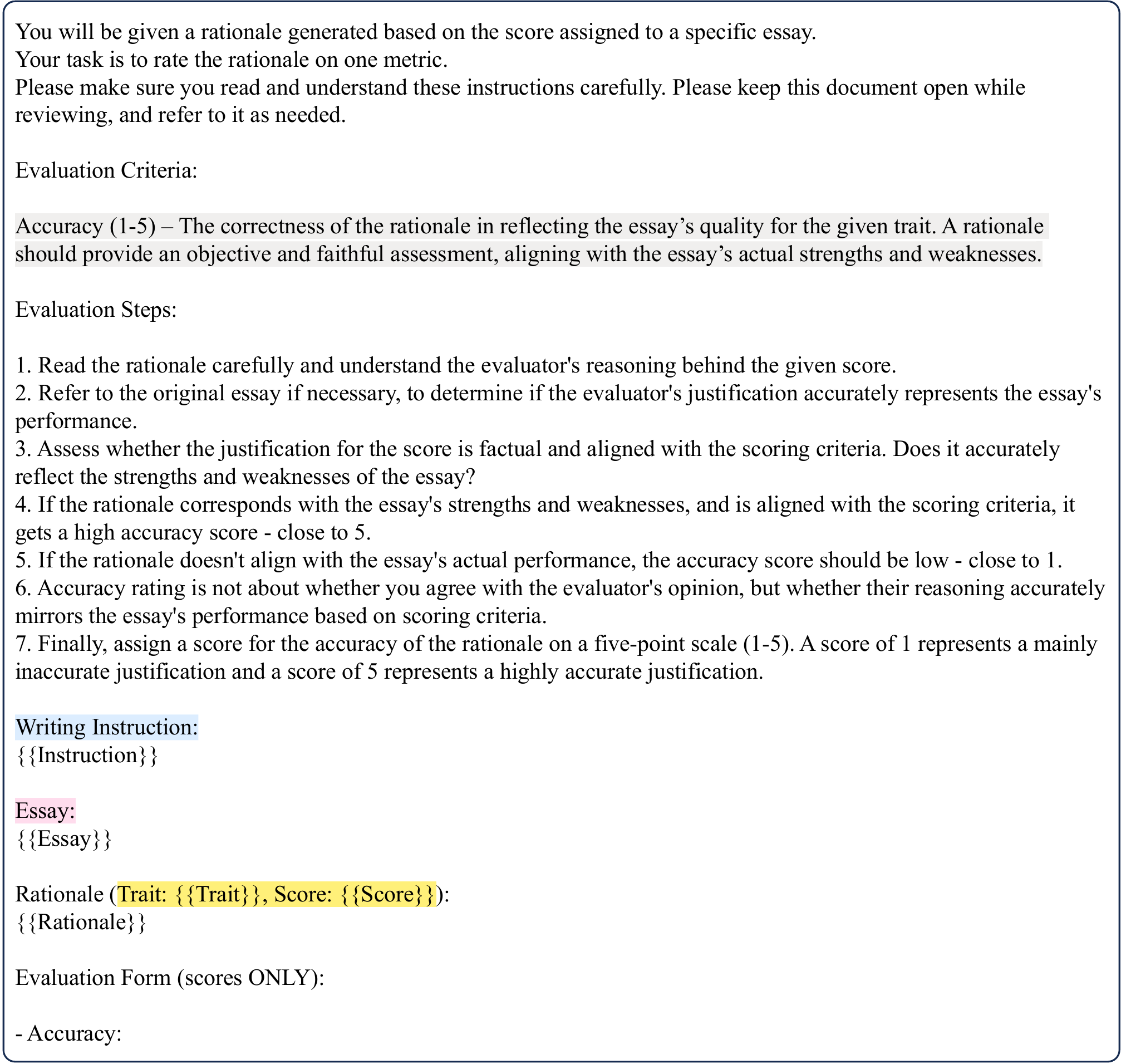}
\caption{Evaluation for \textit{Accuracy}.}
\label{fig: geval-accuracy}
\end{figure*}

\begin{figure*}
\centering
\includegraphics[width=0.95\linewidth]{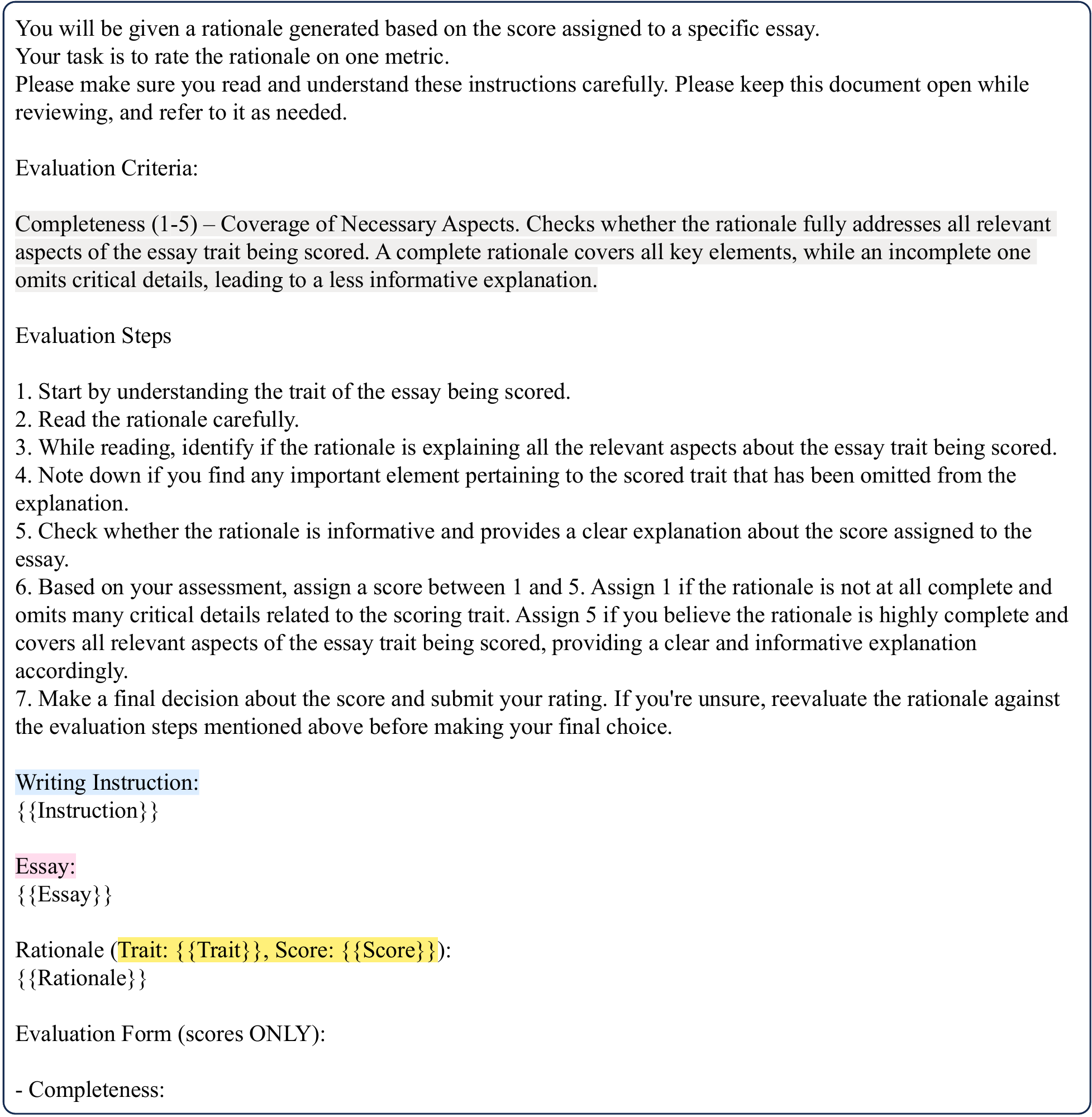}
\caption{Evaluation for \textit{Completeness}.}
\label{fig: geval-completeness}
\end{figure*}

\begin{figure*}
\centering
\includegraphics[width=0.95\linewidth]{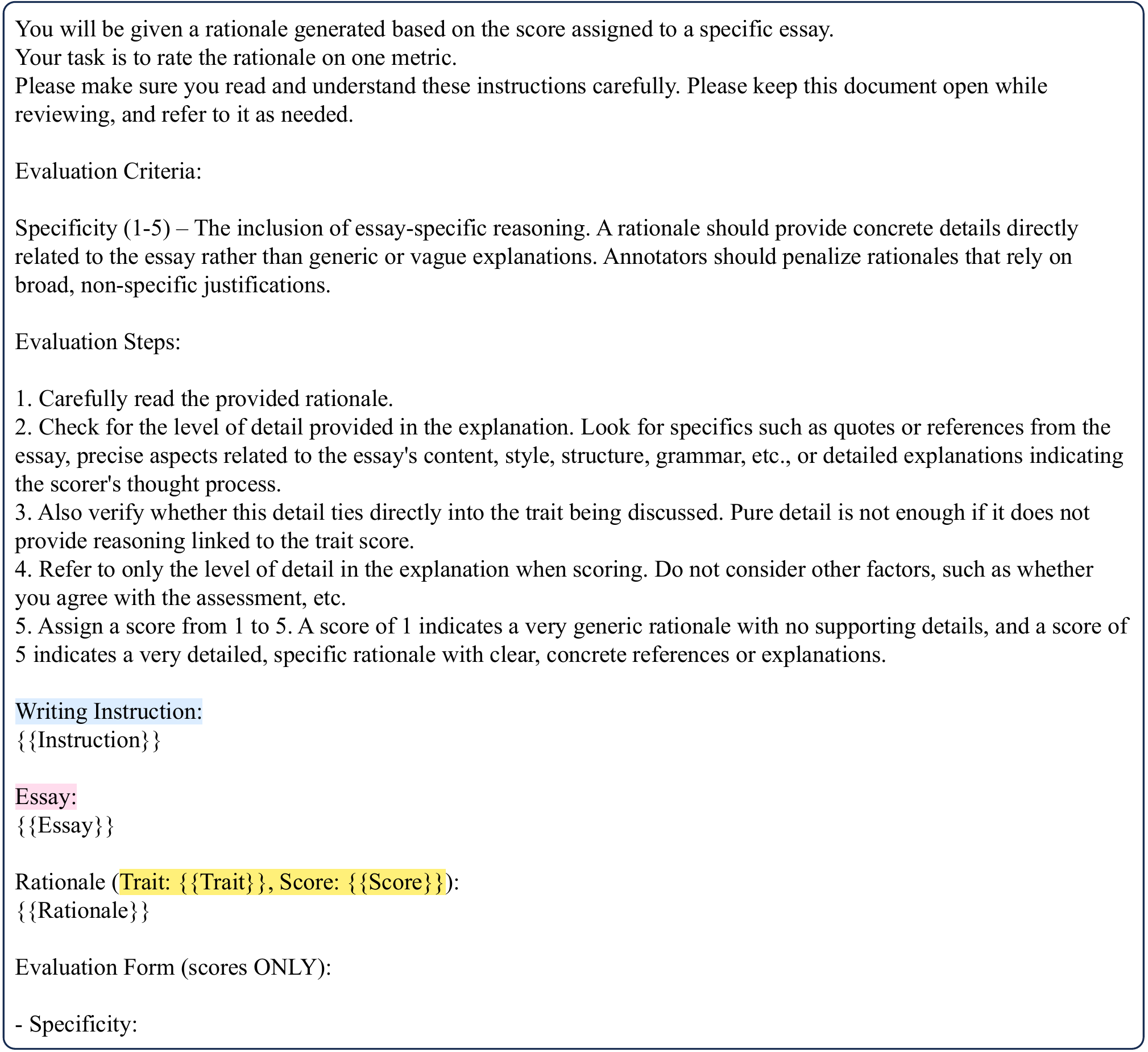}
\caption{Evaluation for \textit{Specificity}.}
\label{fig: geval-specificity}
\end{figure*}

Based on the criteria of each of the four defined dimensions, we generate a CoT for rationale evaluation. We conduct $n$ evaluations using its own steps and then compute a weighted summation to determine the final score: $\sum_{i=1}^{n} p(s_i) \times s_i$, where $p$ denotes the probability of each score and $s$ denotes the score set. The used prompts are illustrated in Figure~\ref{fig: geval-coherence},~\ref{fig: geval-accuracy},~\ref{fig: geval-specificity},~\ref{fig: geval-completeness}.

\end{document}